\pdfoutput=1

\documentclass[11pt]{article}

\usepackage[]{ACL2023}
\usepackage{booktabs}
\usepackage{times}
\usepackage{latexsym}
\usepackage{graphicx} 
\usepackage[T1]{fontenc}

\usepackage[utf8]{inputenc}

\usepackage{microtype}

\usepackage{inconsolata}

\usepackage{hyperref}

\title{AI Coach Assist: An Automated Approach for Call Recommendation in Contact Centers for Agent Coaching}

\author{Md Tahmid Rahman Laskar, Cheng Chen, Xue-Yong Fu, \\ \textbf{Mahsa Azizi}, \textbf{Shashi Bhushan TN}, \textbf{Simon Corston-Oliver} \\
          Dialpad Canada Inc. \\
   Vancouver, BC, Canada \\ 
  \texttt{\{tahmid.rahman,cchen,xue-yong\}@dialpad.com}\\
  \texttt{\{mahsa.azizi,sbhushan,scorston-oliver\}@dialpad.com}}

\begin{document}
\maketitle
\begin{abstract}
In recent years, the utilization of Artificial Intelligence (AI) in the contact center industry is on the rise. One area where AI can have a significant impact is in the coaching of contact center agents. By analyzing call transcripts using Natural Language Processing (NLP) techniques, it would be possible to quickly determine which calls are most relevant for coaching purposes. In this paper, we present ``AI Coach Assist'', which leverages the pre-trained transformer-based language models to determine whether a given call is coachable or not based on the quality assurance (QA) questions asked by the contact center managers or supervisors. The system was trained and evaluated on a large dataset collected from real-world contact centers and provides an effective way to recommend calls to the contact center managers that are more likely to contain coachable moments. Our experimental findings demonstrate the potential of AI Coach Assist to improve the coaching process, resulting in enhancing the performance of contact center agents. 
\end{abstract}

\section{Introduction}
AI has the potential to revolutionize many industries, including the contact center industry. With the growing demand for high-quality customer service, contact centers are constantly seeking ways to improve their processes and enhance their agents' performance. One way to achieve this goal is by providing effective coaching and feedback to agents, which can help them identify areas of improvement and develop the necessary skills to provide exceptional customer service. 
As a common practice, contact center managers or supervisors manually select call recordings to listen in, and grade agents' performance using a rubric that contains questions such as ``\textit{did the agent greet the customer by name}" or ``\textit{did the agent properly resolve the customer issue}" to score the call in order to verify if the agent is following the company’s preferred protocol. The grades given by the managers along with their comments are then shared with the agents to improve their performance. 
However, with the large volume of calls that contact centers receive, it is very challenging for managers or supervisors to determine which calls are most important for agent coaching. Thus, the traditional approaches to randomly select calls for agent coaching has the following limitations: 

\begin{itemize}
\item \textbf{Time-consuming process:} Coaching agents can be a time-consuming process, particularly for managers and supervisors who must manually review large numbers of calls to identify which calls are most relevant for coaching.
\item \textbf{Inefficient use of resources:} Without an efficient and effective process for determining which calls are most relevant for coaching, resources may be wasted on calls that are not critical for improving agent performance.
\end{itemize}

This is where NLP could be useful. By analyzing call transcripts using NLP models, it could be possible to recommend calls to the contact center managers/supervisors that are most relevant for coaching purposes. This will lead to an improved coaching experience by prioritizing the calls for analysis that are more likely to contain coachable moments, resulting in saving time for the contact center managers as well as improving agent performance, ultimately leading to better customer satisfaction. For the purpose of improving real-world contact centers, we present the \texttt{AI Coach Assist} system to assist contact center managers or supervisors by suggesting calls that could be more useful for agent coaching.

In this paper, we explore the concept of our proposed AI Coach Assist system, which leverages the advantage of fine-tuning a pre-trained transformer-based language model \cite{devlin2019bert, DBLP:journals/corr/abs-1910-01108,liu2019roberta, lan2019albert,DBLP:conf/aaai/ZhongLX0022}. 
Moreover, we provide a detailed overview of its development process (implementation and preparation of a balanced dataset to avoid biases), as well as our experimental findings. In addition, we demonstrate how it could be productionized in real-world contact centers to assist managers/supervisors. Note that our model does not automate the scoring of employee performance or replace human review. Instead, our model is intended to help contact center supervisors by recommending calls for coaching their employees instead of the traditional random sampling of calls. 


\section{Related Work}

The significant performance gain achieved via leveraging transformer-based language models \cite{DBLP:conf/nips/VaswaniSPUJGKP17, devlin2019bert, liu2019roberta, lan2019albert} in a wide range of NLP tasks in recent years has also led to the use of transformer-based models in the contact center industry \cite{laskar2022auto,laskar-etal-2022-blink, laskar2022improving, khasanova-etal-2022-developing}. The successful deployment of these models in industries has helped many organizations to enhance their processes, resulting in improved customer satisfaction. In recent years, several studies \cite{fu2022entity,fu-etal-2022-effective-performant} have explored the potential of AI-powered call analysis  (e.g., entity recognition, sentiment analysis, etc.), along with providing real-time assistance to contact center agents. 

In addition to these studies, several commercial solutions have been developed that offer AI-powered call analysis and AI assistance for agents in contact centers. Some of these solutions also offer real-time feedback to agents during calls\footnote{\url{https://cloud.google.com/solutions/contact-center}, accessed in Feb 2023.}\footnote{\url{https://cresta.com/product/agent-assist/}, accessed in Feb 2023.}\footnote{\url{https://www.five9.com/products/capabilities/agent-assist}, accessed in Feb 2023}, allowing them to adjust their behavior and improve their performance in real-time. However, to the best of our knowledge, there is no prior commercial application that assists contact center managers by suggesting calls that could be the most useful to coach agents.

One potential approach for this purpose could be the use of automatic call recommendation, where calls are analyzed using NLP techniques and suggested to the contact center managers based on various factors, such as agents' behavior, issue resolution, customer satisfaction, sales success, etc. These suggested calls can then be analyzed by the managers for coaching purposes to provide relevant feedback to agents. In this regard, we propose \textit{AI Coach Assist}, a system that leverages the transformer architecture to effectively analyze the full call transcripts in contact centers and recommends contact center managers with calls that are more likely to contain coachable moments for a given query. In the following section, we describe how we construct a dataset, which we denote as \textit{QA Scorecard}, to train and evaluate our proposed AI Coach Assist system. 



\begin{table*}
\centering
\small
\begin{tabular}{cccccc}
\toprule
\textbf{Split} & \textbf{Total Samples} &  \textbf{Not Coachable} & \textbf{Coachable} &  \textbf{Avg. Question Length}  & \textbf{Avg. Transcript Length} \\  \midrule
Training & 12065 & 6521 & 5544  & 9.77 & 659.53 \\ \midrule
Validation & 1653 & 891 & 762  & 9.62 & 664.55 \\ \midrule
Test & 3435 & 1855 & 1580  & 9.77 & 727.77 \\ 
\bottomrule
\end{tabular}
\caption{Data distribution on each split (train/valid/test) based on the total number of question-transcript pairs, \textit{coachable} and \textit{not coachable} labels, and the average length of questions and transcripts.}
\label{table:dataset_stat}
\end{table*}

 \begin{table*}
\centering
\small
\begin{tabular}{ll}
\toprule
\textbf{Question Type} & \textbf{Example Question}  \\  \midrule
Account Verification & \textit{Did the agent verify the customer's email address?} \\ \midrule
Addressing Customer & \textit{Did the agent use the customer's name appropriately?} \\  \midrule
Behavioral & \textit{Did the agent show proper empathy statements?} \\ \midrule
Closing & \textit{Did the agent properly end the call?}  \\  \midrule
Providing Complete Information & \textit{Did the agent mention the payment terms in detail?}  \\  \midrule
Customer Identification & \textit{Did the agent verify the customer's information?} \\  \midrule
Customer Satisfaction & \textit{Was the customer happy?} \\  \midrule
Greeting & \textit{Did the agent properly greet the customer?}  \\  \midrule
Information Collection &  \textit{Did the agent collect all necessary information from the customer?} \\  \midrule
Issue Identification &  \textit{Could the agent properly identify the issue?} \\  \midrule
Issue Resolution &  \textit{Could the agent resolve the issue?} \\  
\bottomrule
\end{tabular}
\caption{Example Questions based on Question Types}

\label{table:dataset_distribution_by_type}
\end{table*}

\section{The QA Scorecard Dataset}

We collected our data from real-world contact centers. The dataset consists of customer-agent call conversation transcripts generated using Automatic Speech Recognition (ASR) systems, along with annotations indicating whether a call is coachable or not. The process of annotating the dataset was carefully designed and implemented, as the annotations were performed by real-world contact center managers and supervisors who analyzed the whole conversation/transcript. In this way, we ensure the high quality of the dataset. 

The data annotation works as follows, the managers/supervisors assign a score to the call based on the performance of the agent for a particular question. We consider a call as coachable for a particular question if the call achieves less than 50\% scores, otherwise, we consider the call for that particular question as not coachable. The dataset was collected over a period of one year and includes a diverse range of call types from different industries, with a variety of customer interactions, reflecting the real-world complexities of the contact center industry. The resulting dataset consists of a large number of call transcripts and annotations, providing a robust representation of real-world customer-agent interactions.

 Note that a total of 58 questions are curated, which are distributed among training, validation, and test sets. While constructing the training, validation, and test splits, we observe that the class distribution (whether coachable or not coachable) for many question-transcript pairs was imbalanced. Thus, to ensure an unbiased dataset (as well as to avoid model overfitting), for each question, we ensured that the ratio between \textit{coachable} and \textit{not coachable} classes (or vice-versa) to be at most 1:2.  In Table \ref{table:dataset_stat}, we describe the distribution of our dataset based on our training, validation, and test set.  Meanwhile, to evaluate the performance of \textit{AI Coach Assist} based on the type of the questions, we also categorize the questions into 11 types using human annotators. We show the question types with example questions for each type in Table \ref{table:dataset_distribution_by_type}.

\section{Our Proposed Approach}

\begin{figure*}
  \centering
  \includegraphics[width=\linewidth]{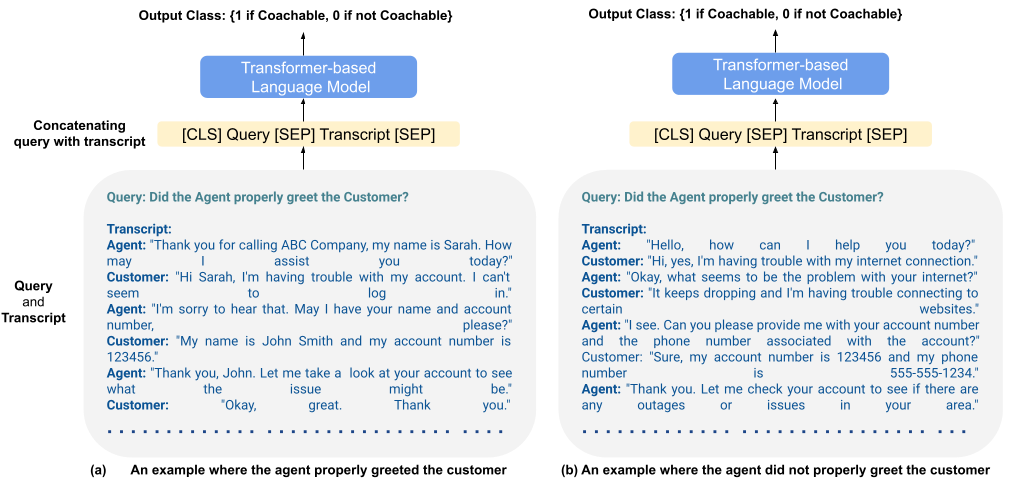} 
  \caption{An overview of our proposed AI Coach Assist model. Given a query and a transcript, the transformer-based language model will determine whether a call is coachable or not. For the given query: "Did the agent properly greet the customer?", on the left (a), we show an example transcript where the agent did proper greeting, i.e., mentioned his/her name as well as the company name. On the right (b), we show an example transcript where the agent did not properly greet the customer as the agent name and the company name were not mentioned.}
  \label{fig:example} 
\end{figure*}

We treat the AI Coach Assist model as a text classification model that combines the query/question given by the contact center manager or supervisor with the call transcript to predict whether a given call is coachable or not. Due to the recent success of fine-tuning pre-trained transformer models for text classification~\cite{devlin2019bert,liu2019roberta,lan2019albert}, we also leverage the pre-trained language models based on the transformer architecture for this task.

As we are doing text classification instead of generation, we give input data to the pre-trained language model as follows (see Figure \ref{fig:example}): at first, we create a text sequence by concatenating the question and the call transcript. Then, this concatenated text sequence is given as input to the language model to learn the contextual relationship between the sentences. The pre-trained transformer language model is fine-tuned to output a probability score for each input sequence, indicating the likelihood that the call is coachable or not, for the given question. Whether a question-transcript pair is \textit{coachable} or \textit{not coachable} is determined based on the probability score of the class having the higher score. 

Since our objective is to build the AI Coach Assist system for real-world contact centers, we consider the following two cases while selecting the pre-trained language models: 

    
\paragraph{(i) Utilize a model to ensure high efficiency:} We choose DistilBERT \cite{DBLP:journals/corr/abs-1910-01108} for this scenario. DistilBERT is a distilled version of BERT \cite{devlin2019bert}, designed to be smaller and faster while retaining a similar level of performance. Despite its smaller size, DistilBERT has been shown to perform similarly to BERT on many NLP tasks, making it a suitable alternative for many NLP applications. This makes it a popular choice for real-world scenarios where computational resources are limited but the preference is to deploy a fast and optimize model in production. 

\paragraph{(ii) Utilize a model to ensure higher accuracy:} For this purpose, we leverage the DialogLED model~\cite{DBLP:conf/aaai/ZhongLX0022}, which was pre-trained on long dialog conversations, having more similarities with our customer-agent conversation dataset. Though in comparison to DistilBERT, the DialogLED model may require higher computational resources for production deployment, it fulfills our criteria of using a model that may provide higher accuracy for being pre-trained on long dialog conversations, mimicking the customer-agent conversations in the real world. In addition, DialogLED can also process long text sequences, contrary to the 512-token limit of most transformer-based models \cite{devlin2019bert,DBLP:journals/corr/abs-1910-01108,liu2019roberta,lan2019albert}. This makes DialogLED a suitable choice to build the AI Coach Assist system since the average length of the transcripts in our QA scorecard dataset is longer than 512 words. 


\section{Experiments}

In this section, we first present the experimental settings and the implementation details of our proposed model. Then we discuss our experimental findings in detail. 

\subsection{Implementation} For the DialogLED model, we adopt the DialogLED-base\footnote{\url{https://huggingface.co/MingZhong/DialogLED-base-16384}} model from the HuggingFace library \cite{wolf2019huggingface}. Specifically, we used the \textit{LEDForSequenceClassification} which adds a classification head on top of the LED (Longformer-Encoder-Decoder) model~\cite{Beltagy2020Longformer}. We ran our experiments in GCP\footnote{\url{https://console.cloud.google.com/}} on an \textit{n1-standard-32} machine with 4 \textit{Nvidia T4} GPUs. A total of $3$ epochs were run, with the training batch size set to $2$\footnote{Larger batch size leads to \textit{Out of GPU Memory} errors.}, and the maximum sequence length set to $1024$. The learning rate was set to $2e-5$. For the DistilBERT model, we leverage its base model from HuggingFace\footnote{\url{https://huggingface.co/distilbert-base-cased}}. We also set the learning rate for DistilBERT to $2e-5$ and ran 3 epochs with the training batch size set to 16 while the maximum sequence length set to $512$. Note that for both models, these hyperparameters were tuned based on the performance in the validation set. The best-performing method in the validation set was then used for evaluation on the test set. 

\subsection{Results \& Discussions}

In this section, we first present the results of our base models. Then we conduct some ablation tests and also compare our proposed models with some classical machine learning baselines to further validate the effectiveness of our approach. Finally, we study the advantages and limitations of our model based on various question types. 
 \begin{table}
\centering
\small
\begin{tabular}{ccccc}
\toprule 
\textbf{Model} & \textbf{Precision} & \textbf{Recall} & \textbf{F1} & \textbf{Accuracy}  \\  \midrule
DialogLED & 67.92 & 63.72 & 65.76& 70.52\\ \midrule
DistilBERT & 62.53 & 58.39 & 60.39 & 66.25 \\ 
\bottomrule
\end{tabular}
\caption{Performance Comparisons between the AI Coach Assist models on our QA Scorecard dataset.}
\label{table:performance_comparisons}
\end{table}


 \begin{table}
\centering
\small
\begin{tabular}{lccc}
\toprule 
\textbf{Model} & \textbf{Precision} & \textbf{Accuracy}  \\  \midrule
DialogLED & 67.92 & 70.52 \\
\textit{ - without query} & 57.76 & 62.83 \\
\textit{ - reduced sequence length = 512}& 66.01  & 67.52 \\
\textit{ - reduced sequence length = 256}& 63.15  & 63.64 \\ \midrule 
DistilBERT & 62.53 & 66.25 \\ 
\textit{ - without query}& 59.32  & 61.22 \\
\bottomrule
\end{tabular}
\caption{Ablation Tests on the QA Scorecard dataset.}
\label{table:ablation_studies}
\end{table}

\subsubsection{Performance of the Base Models}
In this section, we compare the performance of using DialogLED and DistilBERT as the base model for the AI Coach Assist system. Though we consider \textit{precision} and \textit{accuracy} as the main criteria for the production deployment of this system, for this performance evaluation we also consider \textit{recall} and \textit{f1} in addition to \textit{precision} and \textit{accuracy}. 

We observe from our results given in Table \ref{table:performance_comparisons} that the DialogLED model outperforms its counterpart DistilBERT model in terms of all metrics (\textit{precision, recall, f1, and accuracy}). The DialogLED-based model also ensures scores above 60 in all 4 metrics. Moreover, in terms of accuracy and f1, it achieves a score of 70.52 and 65.76, respectively. Meanwhile, both models achieve comparatively lower recall scores, noticeably the DistilBERT model achieves a recall score even below 60. However, in our criteria for production deployment, a highly precise model is more important, with both DialogLED and DistilBERT achieving higher precision scores (67.92 and 62.53, respectively) in comparison to their recall scores (63.72 and 58.39, respectively). 

The superior performance using DialogLED over DistilBERT in all these metrics demonstrates the effectiveness of fine-tuning a language model for contact center telephone transcripts that is pre-trained on dialog conversations. Moreover, since customer-agent conversations can also be quite long and may not fit within the 512 tokens limit of DistilBERT-like models (as shown in Table \ref{table:dataset_stat}), the ability of DialogLED to process input text of larger size may also help it to achieve better performance. In the following section, we conduct some ablation studies to further investigate the effectiveness of our models.


 \begin{table}[t!]
\centering
\small
\begin{tabular}{ccc}
\toprule 
\textbf{Model} & \textbf{Precision} & \textbf{Accuracy} \\  \midrule
TF-IDF + SVM & 57.9 & 57.7 \\ \midrule
TF-IDF + Decision Tree  & 58.0 & 60.8 \\ \midrule
TF-IDF + Random Forest & 59.3 & 60.1 \\ \midrule
TF-IDF + Naïve Bayes & 52.5 & 53.3  \\ \midrule
DialogLED & 67.9 & 70.5 \\ \midrule
DistilBERT & 62.5 & 66.3 \\ 
\bottomrule
\end{tabular}
\caption{Performance Comparisons between some baselines and proposed models on the QA Scorecard dataset.} 
\label{table:baseline}
\end{table}
          
\begin{figure*}
  \centering
  \includegraphics[width=\linewidth]{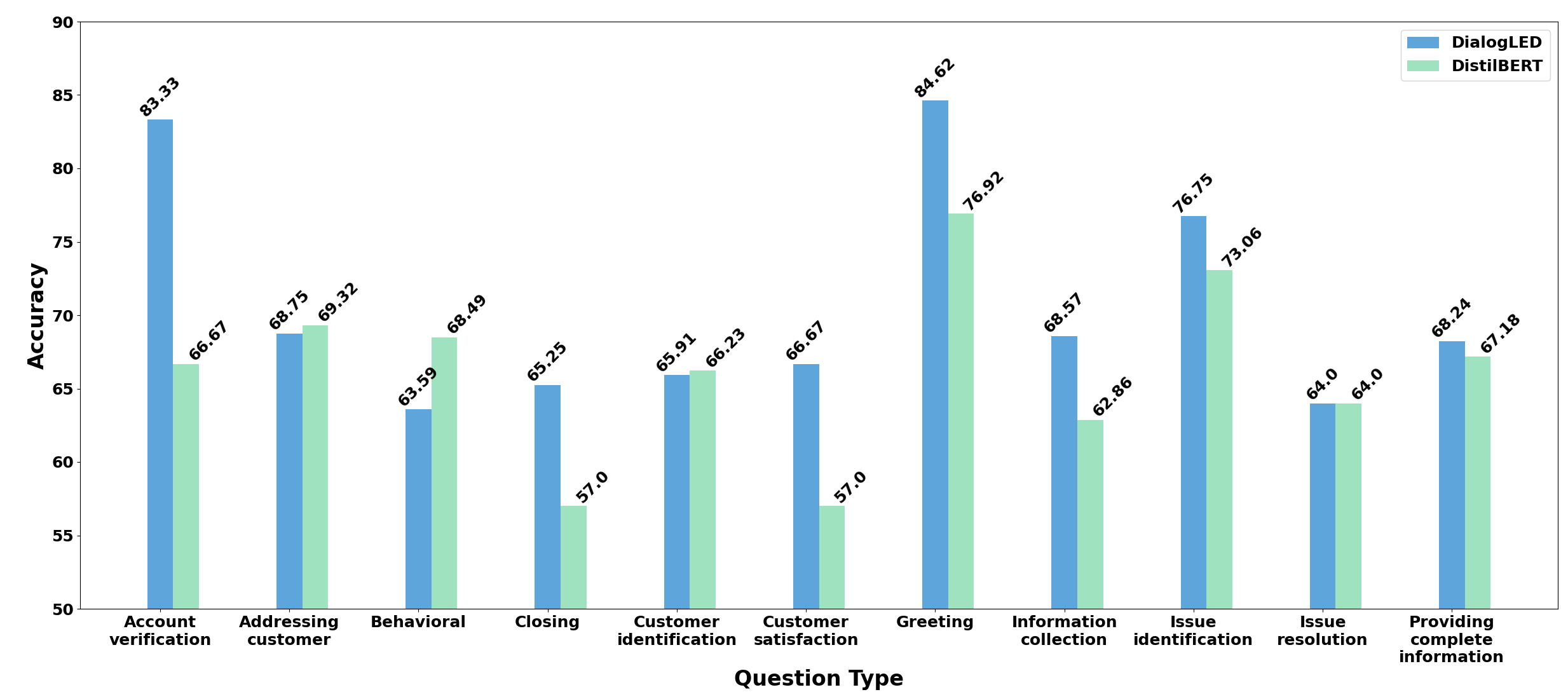} 
  \caption{Performance of DialogLED and DistilBERT on our QA Scorecard dataset based on each Question Type.}
  \label{fig:plot} 
\end{figure*}

\subsubsection{Ablation Studies}

In this section, we conduct some ablation studies to investigate our approach of concatenating the query and the transcript as input for our transformer-based language models (DialogLED/DistilBERT), as well as how the sequence length impacts the overall performance of DialogLED.  We show the results from our ablation study in Table~\ref{table:ablation_studies}.

For our first ablation test, we remove the query from the input text to better study the relationship between the query and the transcript. We
  find that for both models the accuracy is dropped by a great margin if the query is removed. The removal of the query from the input text leads to an accuracy drop of 10.90\% for DialogLED and 8.03\% for DistilBERT. In terms of precision, the performance is deteriorated by 14.96\% and 5.13\%, for DialogLED and DistilBERT, respectively. These findings demonstrate that the model learns to predict the \textit{coachable} and \textit{not coachable} moments in transcripts for the given query based on the concatenated representation of the query and the transcript. 

For our other ablation test, we reduce the input sequence length from our DialogLED model. We find that reducing the input sequence length from 1024 to 512 and 256 leads to a huge drop in accuracy (dropped by 4.71\% and 9.76\%, respectively) and precision (dropped by 2.81\% and 7.02\%, respectively). This demonstrates the effectiveness of using the DialogLED model which can process longer input sequences. 

Moreover, we observe that when the size of the input sequence length for DialogLED is 512 (same as DistilBERT), it still outperforms DistilBERT in terms of both accuracy and precision. This further gives an implication that the utilization of a model that is pre-trained on conversational data is more helpful to improve the performance of the Ai Coach Assist system. 

\subsubsection{Performance against other Baselines}

In this section, we compare our proposed models for the AI Coach Assist system: \textit{DialogLED} and \textit{DistilBERT}, with some baseline models to further study their effectiveness. Below, we describe the baseline models that we use for comparisons:

 \textbf{TF-IDF with Classical Machine Learning Models as Baselines:} We use TF-IDF as keyword-based features for some classical machine learning models, such as Support Vector Machine (SVM) \cite{hearst1998svm}, Random Forest \cite{ho1995randomforest}, Decision Tree \cite{rokach2005decisiontree}, and Naïve Bayes \cite{webb2010naivebayes}, as our baseline models for comparisons. We show our experimental results in Table \ref{table:baseline} to observe that both of our proposed models (the DialogLED model which obtains the highest accuracy and the DistilBERT model which ensures high efficiency) for AI Coach Assist outperform all TF-IDF feature-based classical machine learning approaches. On Average, the DistilBERT model and the DialogLED model outperform the baseline models by 8.97\% and 12.48\% in terms of precision, while 16.20\% and 17.78\% in terms of accuracy, respectively. 





\subsubsection{Performance based on Question Types}

In this section, we conduct an in-depth analysis of the proposed models for AI Coach Assist:  DialogLED and DistilBERT. For our analysis, we investigate their performance in different question types. In Figure \ref{fig:plot}, we show their accuracy on each question type. We observe that for most question types, DialogLED outperforms DistilBERT (the only exceptions are the following question types: \textit{Addressing Customer}, \textit{Behavioural}, and \textit{Customer Identification}). Among the questions where DialogLED outperforms DistilBERT (7 out of 11), the highest performance gains are in question types that are of \textit{Greeting} and \textit{Account Verification}. For \textit{Greeting}, it achieves the best accuracy with a score of 84.62, while for \textit{Account Verification}, the accuracy is 83.33. Meanwhile, even though the DialogLED model achieves an accuracy of at least 60 for all question types, the DistilBERT model achieves quite low scores for some question types  (e.g., only 57\% scores for \textit{Closing} type and \textit{Customer Satisfaction} type questions). For the DialogLED model, it finds the \textit{Behavioral} and the \textit{Issue Resolution} type questions most challenging, as its accuracy drops below 70. Among these two question types, the \textit{Behavioral}  question type achieves the lowest accuracy score of 63.59, followed by \textit{Issue Resolution}, with an accuracy of 64.0. 

\section{Usage in Real World Contact Centers}

In this section, we discuss how the {AI Coach Assist} system can be used in real-world contact centers. Since determining the calls that are coachable is not required in real-time, rather they are required after the call is over, the inference speed of the model may not be an issue in this regard. Moreover, for contact centers where the computing resource is not a problem, our DialogLED-based model could be used, as it achieves better accuracy than its DistilBERT counterpart. Since the size of the trained DialogLED model is 648 MB, the DistilBERT model which takes only 263 MB could be used in scenarios where the computing resource is limited. 

We also prototype our proposed system for usage in a real contact center. Since directly predicting a score to a call might impact the evaluation of agent performance in contact centers, as such metrics could be used by managers for performance evaluation of agents, in our prototype we rather recommend a list of calls to the managers that are highly likely to contain coachable moments for a particular question type. Thus, instead of using those calls for a direct performance evaluation of agents, the managers still require to listen to the conversation or read the ASR-generated transcript. More particularly, using our proposed {AI Coach Assist}, we help the managers with a list of calls that they may use to manually grade agent performance, contrary to the existing methods of random call selection. In this way, the proposed prototype of {AI Coach Assist} may not cause any ethical concerns. 

\section{Conclusion} In this paper, we presented \textit{AI Coach Assist}, a transformer-based pairwise sentence classification model that combines the query/question given by the contact center manager or supervisor with the call transcript to determine which calls are most relevant for coaching purposes. The evaluation results demonstrate the potential of AI Coach Assist to transform the way contact centers coach their agents, providing an efficient and effective method that recommends calls that are the most relevant for coaching purposes. This will help to improve the coaching process and enhance the performance of contact center agents, leading to better customer satisfaction. Note that our model is intended to help contact center supervisors to be more effective in coaching their employees by improving over the random sampling of calls. The model does not automate the evaluation of employee performance by replacing human review. 
 
In the future, we will study how to improve the performance on the question types where the model performs poorly. We will also study how to utilize other question-answering models  \cite{laskar2020contextualized, laskar-etal-2022-domain} or leverage generative large language models \cite{openai2023gpt,anil2023palm2} that can point out the reason for a call being \textit{coachable} and \textit{not coachable}. 

\section*{Limitations}
As our models are trained on customer-agent conversations in English, they might not be suitable to be used in other domains, types
of inputs (i.e written text), or languages. Moreover, as we demonstrated in the paper that the model has limitations in certain question types, the user needs to decide which question types to be used when deploying the system in production. Though the DialogLED model performs better, it requires higher computing resources. On the contrary, even though the DistilBERT model consumes lower memory, its performance is poorer than the DialogLED model. 

\section*{Ethics Statement}
\begin{itemize}
    \item \textbf{Data Annotation:} Since the calls are annotated by real-world contact center managers/supervisors, we did not require any additional compensation for this annotation. Rather, we develop a system where the managers/supervisors put their scores for different call conversations in their contact centers.  To map the questions to different question types, Labelbox\footnote{\url{https://labelbox.com/}} was used for data annotation and the annotators were provided with adequate compensation (above minimum wages). 

\item \textbf{Privacy:} There is a data retention policy available so that the call transcripts will not be used if the user does not give permission to use their call conversations for model training and evaluation. To protect user privacy, sensitive data such as personally identifiable information (e.g., credit card number, phone number) were removed while collecting the data. 

\item \textbf{Intended Use by Customers:} Note that our model is intended to help contact center supervisors to be more effective in coaching their employees by improving over the random sampling of calls. The model does not automate the scoring of employee performance or replace human review. 

\item \textbf{Prevention of Potential Misuses:} Below, we discuss some of the potential misuses of the system and our suggestions to mitigate them:

 \textit{\textbf{(i) Automatic Performance Reviews of Agents by Considering all Recommended Calls as Bad Calls:}} One potential misuse of the system could be the evaluation of agent performance by considering all recommended calls as bad calls without any manual review of the call. To mitigate this, we can do the following: 
 \begin{itemize}
     \item Contact center supervisors that use this system must be properly instructed that this system does not determine whether an agent performs badly in a certain call. Rather, the intention of the proposed system is to only suggest a set of calls to the managers (instead of randomly selecting calls) that they need to manually review to determine whether the agent requires coaching or not.

 \end{itemize}

  \textit{\textbf{(ii) Considering Agents with More Recommended Calls as an Indicator to Poorer Agent Performance:}} Another potential misuse of the system is if contact center managers start considering that if more calls are recommended by our system for a particular agent, then the agent is more likely to perform poorly. To prevent this, we can do the following: 
 \begin{itemize}

     \item  We may suggest some positive calls as well as negative calls to the managers. Here, positive calls are the ones that our system rates with a very high score and categorizes as not requiring any coaching. Whereas negative calls are the ones that our system rates with quite lower scores and classifies as coaching required. To avoid any misuse of the suggested calls, the proposed AI Coach Assist system should never reveal to the managers whether a call requires coaching or not. Rather it should only allow the managers to make the final decision on whether the call is a positive call or a negative call.  Once the suggested calls are manually reviewed by the managers and categorized as positive by them, these calls can then be used to train other agents that require improvement in certain areas, whereas a call categorized as negative can be used to train a particular agent who did not perform well (i.e., requires coaching) in that specific call.
     \item In addition, to avoid suggesting too many calls for the same agent,  the system may suggest  only a certain number of calls (not above a pre-defined limit) per agent to the managers. 
     
 \end{itemize}

   \textit{\textbf{(iii) Using Bad Questions For Model Development:}} In some contact centers, there may be questions that are used for evaluating agent performance which may contain any potential biases toward a specific race or gender. We can prevent this in the following way: 
 \begin{itemize}

     \item The system should only respond to a pre-selected set of questions that were used during the training phase of the model. Any questions that may pose any ethical concerns or potential biases should not be used while training the model such that these  questions can also be automatically ignored during the inference phase. 
 \end{itemize}

\item \textbf{License:} We maintained the licensing requirements accordingly while using different tools (e.g., HuggingFace).
\end{itemize}

\section*{Acknowledgements} We appreciate the reviewers for their excellent review comments that helped us to improve the quality of this paper. We would also like to thank \textbf{Shayna Gardiner} and \textbf{Elena Khasanova} for reviewing the ethical concern of the proposed system. 

\bibliography{anthology,custom}
\bibliographystyle{acl_natbib}

\appendix



\end{document}